\documentclass[10pt,twocolumn,letterpaper]{article}

\usepackage[pagenumbers]{cvpr} 

\definecolor{cvprblue}{rgb}{0.21,0.49,0.74}
\usepackage[pagebackref,breaklinks,colorlinks,allcolors=cvprblue]{hyperref}
\usepackage{algorithm,algorithmic,multirow}
\usepackage{booktabs}
\usepackage{float}
\usepackage{makecell}
\usepackage{textpos}

\def\paperID{18372} 
\def\confName{CVPR}
\def\confYear{2026}

\title{RoboScape-R: Unified Reward-Observation World Models for Generalizable Robotics Training via RL}

\author{
Yinzhou Tang$^*$\textsuperscript{1},
Yu Shang$^*$\textsuperscript{1},
Yinuo Chen$^*$\textsuperscript{1},
Bingwen Wei\textsuperscript{1},
Xin Zhang\textsuperscript{2},
Shu'ang Yu\textsuperscript{1},
Liangzhi Shi\textsuperscript{1},\\
Chao Yu\textsuperscript{1},
Chen Gao\textsuperscript{1},
Wei Wu\textsuperscript{2},
Yong Li$^\dagger$\textsuperscript{1}\\
\textsuperscript{1} Tsinghua University \\
\textsuperscript{2} Manifold AI \\
\textsuperscript{*} Euqal Contribution \\
\textsuperscript{$^\dagger$} liyong07@tsinghua.edu.cn
}

\begin{document}
\maketitle
\begin{abstract}
Achieving generalizable embodied policies remains a key challenge. Traditional policy learning paradigms, including both Imitation Learning (IL) and Reinforcement Learning (RL), struggle to cultivate generalizability across diverse scenarios. While IL policies often overfit to specific expert trajectories, RL suffers from the inherent lack of a unified and general reward signal necessary for effective multi-scene generalization.
We posit that the world model is uniquely capable of serving as a universal environment proxy to address this limitation. However, current world models primarily focus on their ability to predict observations and still rely on task-specific, handcrafted reward functions, thereby failing to provide a truly general training environment.
Toward this problem, we propose RoboScape-R, a framework leveraging the world model to serve as a versatile, general-purpose proxy for the embodied environment within the RL paradigm. We introduce a novel world model-based general reward mechanism that generates ``endogenous'' rewards derived from the model's intrinsic understanding of real-world state transition dynamics.
Extensive experiments demonstrate that RoboScape-R effectively addresses the limitations of traditional RL methods by providing an efficient and general training environment that substantially enhances the generalization capability of embodied policies. Our approach offers critical insights into utilizing the world model as an online training strategy and achieves an average 37.5\% performance improvement over baselines under out-of-domain scenarios.
\end{abstract}    
\section{Introduction}

Developing robust and generalizable embodied policies stands as a critical and persistent challenge toward achieving Artificial General Intelligence (AGI). While continuous advancements in foundational models and training paradigms have empowered embodied policies to achieve extensive manipulation capabilities, their generalization capacity to handle novel environments and transfer to unseen tasks remains a primary limitation.
The prevailing training paradigm, Imitation Learning (IL), often relies on supervised policy optimization guided by manually engineered objectives or large-scale, expert-curated data~\cite{kim2025fine, li2025robotic, fang2019survey}. This dependency frequently leads to overfitting to the specific training scenarios and the expert's optimal trajectories~\cite{celemin2022interactive, hussein2017imitation}. In contrast, Reinforcement Learning (RL), which leverages reward signals~\cite{kaelbling1996reinforcement, li2017deep}, inherently encourages the generation of more diverse exploratory trajectories~\cite{singh2022reinforcement, garaffa2021reinforcement, tang2025deep}. However, RL still contends with fundamental limitations regarding broad generalization across task families~\cite{dulac2019challenges}. Furthermore, scaling RL to a unified, multi-task policy is complicated by the inherent difficulty in designing a universal and consistent reward function applicable across highly heterogeneous environments.

To address this generalization bottleneck, world models—which are learned predictive models able to forecast the next observation based on the current state and action signals~\cite{ding2025understanding, shangsurvey,long2025survey}—offer a compelling alternative.
By providing an accurate, internal simulation environment, world models thus hold the potential to serve as a novel and powerful data-efficient training paradigm for robotics, significantly advancing the acquisition of broadly generalizable embodied skills.
As trained on massive embodied scenarios, world models can learn the transition dynamics between different states and their relationships with actions via unsupervised methods, thus having been regarded as a general environmental proxy~\cite{yang2023unisim,bardes2023v,assran2025v}.

Despite their predictive power, two primary technical barriers currently impede the seamless integration of existing world models as applicable RL environments: the lack of a generalizable reward signal and insufficient robust action controllability. First, current world models are typically designed only to predict future observations based on historical states and actions. This design inherently omits the explicit reward and termination signals required for policy optimization. Second, utilizing a world model as a training environment imposes rigorous demands on action controllability, which ensures the world model can synthesize consistent and physically plausible observations, even when processing challenging or out-of-distribution control inputs. While some contemporary research explores deploying world models for RL, these approaches often resort to simplistic prediction heads~\cite{wu2024ivideogpt} or external reward models~\cite{chandra2025diwa} to synthesize necessary signals. This reliance on ``exogenous'' reward systems forces the policy to optimize against an artificially prescribed curve, consequently inheriting the well-known limitations of extrinsic rewards regarding generalization to novel tasks and scenarios.

In this work, we introduce RoboScape-R, a novel RL framework that deploys a world model as the primary simulation environment for generalizable embodied policy training.
To realize this, we meticulously design a dual-world model pipeline specialized in processing agent actions and concurrently producing instruction-following observation transitions.
Besides, we develop an intrinsic, universal reward signal derived directly from the world model itself. 
This universal reward capability significantly enhances multi-scenario policy training, which is pivotal for achieving out-of-distribution generalization. 
Specifically, our contributions can be divided into the following three points:
\begin{itemize}
    \item We establish a world model-derived general reward mechanism that generates universal reward signals across heterogeneous tasks, allowing us to train policies capable of broad generalization.
    \item We present a pioneering world model-centric RL methodology that integrates the world model as a versatile environment simulator, which provides all essential environment signals enabling the efficient training of generalizable embodied policies.
    \item Empirical validation confirms the efficacy of our framework, showing that policies trained within the RoboScape-R environment achieve a 37.5\% performance improvement in the out-of-domain scenarios.
\end{itemize}
\section{Related Works}

\subsection{Embodied World Models}
World models have emerged as a pivotal technical pillar for embodied intelligence to generate the next observation with the control of the historical observation and action. The world models can be categorized into diffusion-based (e.g., Wan~\cite{wan2025wan}, CogVideoX~\cite{yang2024cogvideox}), autoregressive-based (e.g., Genie~\cite{bruce2024genie}, Lumos-1~\cite{yuan2025lumos}), and hybrid models (e.g., NOVA~\cite{deng2024autoregressive}, LongScape~\cite{shang2025longscape}). These models can achieve high fidelity and temporally coherent video generation. Thus, they are utilized as a data generator in the embodied domain to solve the data scarcity problem~\cite{jang2025dreamgen, jiang2025enerverse}. Furthermore, the application of utilizing a world model as an offline policy evaluator is also primarily studied~\cite{shang2025roboscape, li2025worldeval}. However, owing to that current world model can only provide future observations, which limits its application to be utilized as an online environment for policy training. Serving the world model as an online environment for robotics training via RL requires it to provide a reasonable reward, additionally. Although some current works are exploring utilizing the world model as the environment~\cite{chandra2025diwa, jiang2025world4rl}, they use an additional reward model as the reward proxy and directly fit the manually designed reward, which lead to the problem of reward unstablity and limited generalization.

\subsection{RL for Embodied Policy Training}
Reinforcement learning (RL) has become a core method for training robot policies. The traditional training methods for robot policies are mainly based on imitation learning (IL), that is, the paradigm of supervised fine-tuning (SFT), but they severely limit the generalization of policies. Reinforcement learning autonomously optimizes goal-oriented behavior through the interaction and trial and error between the policy and the environment, and can train more robust strategies. SimpleVLA-RL~\cite{li2025simplevla}, which builds on the OpenVLA and GRPO frameworks, has shown that reinforcement learning can enhance the long-horizon planning capabilities of VLA models in data-scarce scenarios. RL4VLA~\cite{liu2025can} conducted empirical assessments of PPO, GRPO, and DPO. VLA-RL~\cite{lu2025vla} put forward a specialized reward model for robotic processes and improved the data processing pipeline. iRe-VLA~\cite{guo2025improving} proposed a framework that alternates between RL exploration and SFT updates. However, most of these methods are optimized based on a rule-based or proxy-based reward, and this exogenous reward paradigm limits the generalization of the strategy when training a policy in multiple tasks.
\begin{figure*}
    \centering
    \includegraphics[width=1\linewidth]{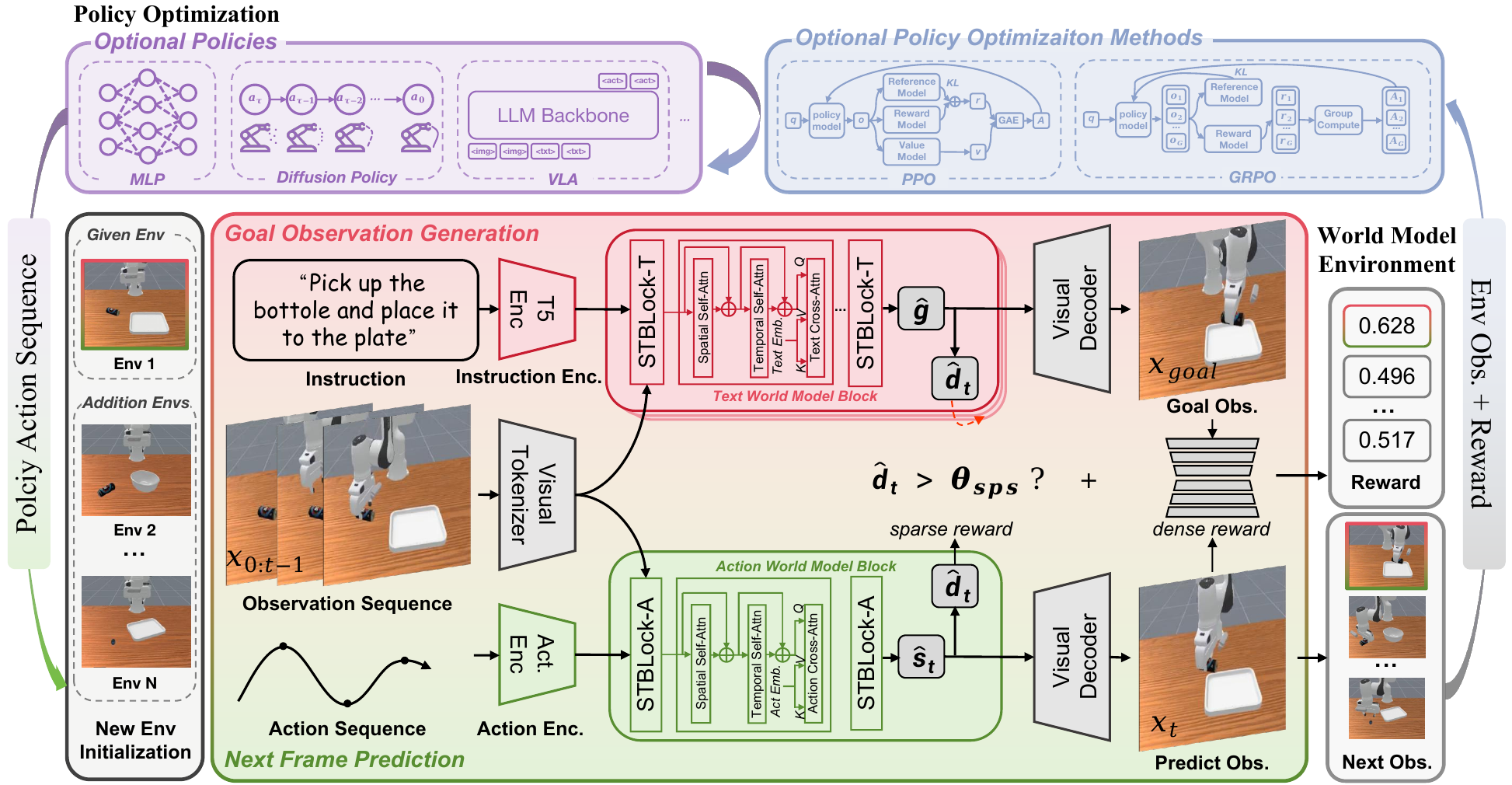}
    \vspace{-0.5cm}
    \caption{Overall structure of the proposed RoboScape-R pipeline. It mainly consists of a World Model-based Environment with General Reward and Policy Optimization. We designed a world model-based general reward to train the policy in multiple environments. The world model environment is a dual-world model structure, in which the action world receives the action and provide predicted observation while the text world model provide reward signal with a generated goal observation. This paradigm allows policy to interact with multiple environments to train a generalizable policy.}
    \label{fig:main}
\end{figure*}

\section{Methods}
The overall structure of the proposed RoboScape-R pipeline mainly includes two parts. World model serves as the environment receiving the predicted action from the policy, and providing the next frame observation and the corresponding reward with generalized and scalable interfaces. The second part is scalable policy options that receive the observation and provide the predicted action. The overall structure of our method is shown in Fig.~\ref{fig:main}. In this Section, we first introduce the architecture of the world model and how to utilize it as the RL environment to provide a general and unified reward, and then we propose a pipeline for using the world model as an environment to train a policy.

\subsection{World Model with General Reward}
\paragraph{Architecture of the World Model}
Existing world models can respond to current actions and provide observations subsequent to action execution. However, they are unable to simultaneously generate rewards to guide the policy during RL training. We propose a world model-based RL environment with an action world model $WM_{act}$ and a text world model $WM_{txt}$ to interact with the policy while providing both observations and rewards. The action world model is responsible for responding to and interacting with the policy actions to provide observations, while the text world model constructs a golden trajectory as a reward prior and generates rewards by incorporating predicted observations.

As shown in Fig.~\ref{fig:main}, the two world models share a similar architecture of an auto-regressive Transformer-based framework but with distinct parameters, which both predict the future observation and the done signal based on historical observation and the control signal. The forward process of the world model can be formulated as follows:
\begin{align}
    \hat{x_{t}}, \hat{d_{t}} = WM\left(x_{0:t-1}, c\right),
\end{align}
in which $x_t \in \mathbf{R}^{H \times W \times 3}$ refers to the observation RGB frame, $\hat{d_t}$ for the predicted done signal, and $c$ refers to the control signal. The control signal is $a_{0:t} \in \mathbf{R}^{d \times t}$ for the action world model, referring to the $d$-dimensional robotic action, and $i$ for the text world model refers to the text instruction.

The architecture of the world model mainly includes the visual tokenizer for observation tokenization, an action or instruction encoder for control information encoding, an action or text-based Spatial-Temporal Block for conditioned next token and done signal prediction, and a visual decoder for predicted token decoding.

Specifically, we utilize MAGVIT-2~\cite{yu2023language} as the visual tokenizer to compress observation $x_{0:t} \in \mathbf{R}^{T \times H \times W \times 3}$ into discrete latent tokens $g_{0:t}$ or $s_{0:t} \in \mathbf{R}^{T \times H' \times W' \times D}$, where $D$ refer to the dimension of the latent tokens and $H' = H / \alpha$, $W' = W / \alpha$ is the downsampled spatial dimension, and $\alpha$ refers to the donwsample factor. Then we design a conditioned Spatial-Temporal Block to predict the next token and the done signal with the control signal, that is, the text instruction or the action sequence. It can be instantiated as the text-conditioned Spatial-Temporal Block STBlock-T and the action-conditioned Spatial-Temporal Block STBlock-A for text and action world models. We use an instruction or action encoder to encode control signals into the embeddings. For text world models, we utilize the T5 encoder~\cite{raffel2020exploring} as the instruction encoder to encode the task instruction $i$ into the text embedding $c_{i}$, and utilize an MLP as the action encoder to encode the 7-DoF robotic action $a$ into the dense action embedding $c_{a}$ in the action world model. Furthermore, we add a cross-attention mechanism into the Spatial-Temporal Block~\cite{bruce2024genie} for control signal injection. 
Thus, we can get the predicted tokens $\hat{g}$ or $\hat{s_t}$ after the stacked Spatial-Temporal Blocks. Furthermore, we also add an MLP to process the current frame to provide the current done signal $\hat{d_t}$. 

To optimize our model, we use a hybrid loss function consisting of a visual loss $\mathcal{L}_{vis}$ loss of tokens for observation prediction and a done loss $\mathcal{L}_{d}$ for done signal prediction. The visual loss is a cross-entropy loss that can be formulated as
\begin{align}
    \mathcal{L}_{vis} = -\sum^{T}_{t=0}z_t\text{log}p(\hat{z_t}),
\end{align}
in which $z_t$ refers to $s_t$ for action world models and $g_t$ for text world models.
Then, we utilize an RMSE loss as the loss for the done signal prediction:
\begin{align}
    \mathcal{L}_d = \text{RMSE}(\hat{d_t}, d_t).
\end{align}
Thus, the total loss can be formulated as follows:
\begin{align}
    \mathcal{L} = \mathcal{L}_{vis} + \mathcal{L}_d.
\end{align} 

\paragraph{World Model-based General Reward}
In order to achieve a stable optimization for training a generalizable policy, we propose a world model-based general reward. It is an endogenous and unsupervised reward instead of directly fitting the manually designed reward function. The reward is composed of a dense reward $\mathcal{R}_{den}$ for fine-grained guidance to accelerate policy convergence and reduce ineffective exploration, and a sparse reward $\mathcal{R}_{sps}$ to enhance generalization and avoid getting stuck in local optima. 

For the dense reward $\mathcal{R}_{den}$, we calculate the similarity between the current observation $x_t$ and the target observation $x_{goal}$ by LPIPS, and it can be formulated as
\begin{align}
    \mathcal{R}_{den} = \text{LPIPS}(x_t, x_{goal}).
\end{align}
The goal observation $x_{goal}$ provides a prior of the final state of the accomplished task, and it is generated by the text world model based on the initial observation of the environment and the text instruction. Specifically, the text world model auto-regressively generates the predicted next frame observation and the done signal while generating the golden trajectory, and we task the first frame when the done signal is greater than the threshold $\theta$ as the predicted goal observation. The process can be formulated as:
\begin{align}
    x_{goal} = \mathbf{Dec}(g_{t^*}), t^*=\text{min}\{t+1|\hat{d_{t}^{txt}} \geq \theta\},
\end{align}
in which $\mathbf{Dec}$ refers to the visual decoder, and $\theta$ refers to the threshold for the done signal for considering the task as fully accomplished.
Furthermore, the sparse reward $\mathcal{R}_{sps}$ is determined by the predicted done signal $\hat{d^{act}_t}$ following
\begin{align}
    \mathcal{R}_{sps} = 1 \text{ if } d^{act}_t \geq \theta_{sps} \text{ else } 0,
\end{align}
in which $\theta_{sps}$ refer to the threshold for sparse reward.
Thus, the complete reward can be formulated as
\begin{align}
    \mathcal{R} = \mathcal{R}_{sps} + \mathcal{R}_{den}.
    \label{eq:reward}
\end{align}

\begin{algorithm}[t]
    \caption{World Model-Based Policy Optimization}
    \label{alg:main}
    \begin{algorithmic}[1]  
        \REQUIRE {Number of environments $N$; given environment $e_0$, text instruction $i$; text world model $WM_{txt}$; action world model ${WM_{act}}$; policy network $\pi_\Theta(a|s)$; similarity function $\text{sim}(\cdot, \cdot)$; total iterations $K$; steps per iteration $T$; done threshold $\theta \in (0,1)$; sparse reward weight $\alpha \in [0,1]$; dense reward weight $\beta = 1-\alpha$; learning rate $\lambda$.}

        \STATE \textbf{Generalizable Environment Initialization}
        \STATE $\mathcal{E} \leftarrow \{e_0\}$
        \FOR{each environment $n \in \{1, \ldots, N\}$}
            \STATE $\mathcal{E} \leftarrow \mathcal{E} \cup \text{Init($e_n$)}$ \COMMENT{Init additional environments}
        \ENDFOR
        \FOR{each environment $n \in \{0, \ldots, N\}$}
            \STATE $x_n \leftarrow x_n^0$, $t \leftarrow 0$, $d \leftarrow 0$ \COMMENT{Initialize current observation, step, and done signal}
            \WHILE{${\hat{d_t^{txt}}} < \theta$} 
            \STATE {$(x_t, d) \leftarrow WM_{txt}(x_{0:t-1}, i)$} 
            \COMMENT{Recursively generate until done}
            \STATE $t \leftarrow t+1$
            \ENDWHILE
            \STATE $x_{goal,n} \leftarrow x_t$ \COMMENT{Final frame for environment $i$}
        \ENDFOR

        \FOR{each environment $n \in \{0, \ldots, N\}$}
            \STATE $x_n \leftarrow x_n^0$ \COMMENT{Reset to initial observation}
        \ENDFOR
        
        \FOR{$k = 1$ to $K$} 
        
            \STATE $\mathcal{D} \leftarrow \emptyset$ 
            \COMMENT{Total training iterations}
            
            \STATE \textbf{Policy Rollout in the Environment}
            \COMMENT{Initialize trajectory buffer}
            \FOR{$t = 1$ to $T$}
                \FOR{each environment $n \in \{0, \ldots, N\}$}
                    \STATE $a_i \leftarrow \pi_\theta(\cdot|s_i)$ \COMMENT{Policy outputs action}
                    \STATE $(x_{n,t}, \mathcal{R}_{sps}) \leftarrow WM_{act}(x_{n,0:t-1}, a_{0:t-1})$ \COMMENT{Action World Model generates next observation and sparse reward}
                    \STATE $\mathcal{R}_{den} \leftarrow \text{sim}(x_{n,t}, x_{goal,n})$ \COMMENT{Dense reward from similarity to Final frame}
                    \STATE $\mathcal{R}_i \leftarrow \alpha \cdot \mathcal{R}_{sps} + \beta \cdot \mathcal{R}_{den}$ \COMMENT{Total reward}
                    \STATE $\mathcal{D} \leftarrow \mathcal{D} \cup \{(x_n, a_n, \mathcal{R}_n)\}$ \COMMENT{Store transition}
                    \STATE $x_n \leftarrow x_t$ \COMMENT{Update current state}
                \ENDFOR
            \ENDFOR
            \STATE \textbf{Policy Optimization}
            \STATE $\Theta \leftarrow \Theta+\lambda \cdot \nabla_{\Theta}J\left(\Theta;\mathcal{D}\right)$ \COMMENT{Update policy with collected trajectories}
        \ENDFOR

        \RETURN $\pi_\Theta$
    \end{algorithmic}
    \label{algo:rl}
\end{algorithm}

\subsection{World Model-based RL Training}
A well-trained world model can serve as a proxy for the real world. Thus, it can be modified to utilize as the environment to train the policy with reinforcement learning by understanding the task and providing a reasonable reward for policy optimization. The paradigm we proposed of using world models as environments mainly includes three parts: Generalizable Environment Initialization, Policy Rollout in the Environment, and Policy Optimization. The pseudo-code for training the RL with the world model as the environment can be found in Algo.~\ref{algo:rl}.

\paragraph{Generalizable Environment Initialization}
During environment initialization, unlike the environment based on physical simulators that only initialize a single type of environment, the world model, through learning from massive embodied scenarios, enable a single model to predict diverse environmental dynamic transitions, covering various embodied scenarios.
Owing to this, the policy can interact with multiple environments simultaneously, enabling the generalizability of the policy. Furthermore, in order to provide general rewards, we also generate the goal observation $x_{goal}$ of the environment. We utilize the text world model to generate the goal observation based on the initial observation and the instruction to provide a prior of the final state of accomplishing the task in the given environment. Specifically, we use the text world model to auto-regressively generate the predicted next frame and the done signal, and we take the first frame when the done signal is greater than the threshold $\theta$ as the predicted final observation, which can be described as follows:

\paragraph{Policy Rollout in the Environment}
Given an action sequence generated by the policy $\pi_{\Theta}$, the action world model is able to respond to the given action and generate the corresponding next observation. This interaction is accomplished in the action world model to generate the next frame observation, which can be formulated as
\begin{align}
    x_t = \mathbf{Dec}(\hat{s_t}|x_{0:t-1}, e_{a, t}).
\end{align}
Simultaneously, we provide a reward for the current timestep by combining a dense reward generated by the similarity between the current observation and the goal observation, and the sparse reward following Eq.~\ref{eq:reward}.

\paragraph{Policy Optimization}
This world model-based environment is a general and flexible paradigm that can support various implementations, including diverse reinforcement learning (RL) optimization algorithms and policies, as long as they comply with the general interfaces for observation, reward, and done signals.
This process can be described as
\begin{align}
    \Theta \leftarrow \Theta+\gamma \cdot \nabla_{\Theta}J\left(\Theta;D\left(X, R\right)\right),
\end{align}
in which $\Theta$ refers to the policy parameters, $\gamma$ refers to the learning rate, $J$ refers to the object function, $D$ refers to the data funciton, $X = \{x_{0:t}^i\}_{i=0}^N$ refers to the collected observation trajectory, and $R = \{ \mathcal{R}^i \}_{i=0}^{N}$ refers to the corresponding rewards.

\begin{table*}[h]
    \centering
    \caption{Overall performance of the success rate in percentage (\%) for different policies trained in different environments with different tasks and policy and optimization method combinations. P.\&P. refers to the Pick and Place, and M.A. refers to Move to Aim. IND refers to the in-domain evaluation, and OOD refers to the out-of-domain evaluation.}
    \label{tab:overall}
    \begin{tabular}{clcccc|cccc}
        \toprule
        \multirow{2}{*}{\centering{Policy}} & \multirow{2}{*}{\centering{Training Methods}} & \multicolumn{4}{c|}{IND} & \multicolumn{4}{c}{OOD} \\
        \cmidrule{3-10}
         &  & P.$\&$P. & Push & Pull & M.A. & P.\&P. & Push & Pull & M.A. \\
        \midrule
        \multirow{3}{*}{MLP} 
        & Supervised Fintune & 12.0 & 84.5 & 57.7 & 35.8 & 3.5 & 13.3 & 10.2 & 10.6 \\
        & RL w. ManiSkill & 85.0 & 98.5 & 64.2 & 87.6 & 32.1 & 15.6 & 15.5 & 42.9\\
        & RL w. World Model & \textbf{86.7} & \textbf{98.7} & \textbf{65.2} & \textbf{88.1} & \textbf{74.2} & \textbf{87.3} & \textbf{51.3} & \textbf{72.0} \\
        \midrule
        \multirow{3}{*}{OpenVLA}
        & Supervised Fintune & 22.5 & 94.4 & 73.5 & 67.5 & 47.5 & 84.0 & 12.5 & 24.5 \\
        & RL w. ManiSkill & 92.5 & 97.7 & 89.1 & 92.5 & 54.5 & 93.0 & 39.6 &41.5 \\
        & RL w. World Model & \textbf{95.5} & \textbf{98.4} & \textbf{91.4} & \textbf{94.6} & \textbf{84.1} & \textbf{97.5} & \textbf{85.8} & \textbf{82.1} \\
        \bottomrule
    \end{tabular}
\end{table*}

\begin{table}[]
    \centering
    \caption{Success rate in percentage (\%) for multi-task training in ManiSkill and the world model. For the pick and place task, we select the plate as the container and select two objects.}
    \label{tab:multi_task}
    \begin{tabular}{l|cc|cc}
        \toprule
        \multirow{2}{*}{Environments} & \multicolumn{2}{c|}{P.\&P.} & \multicolumn{2}{c}{M.A.} \\
        \cmidrule{2-5}
         & pepper & peach & pepper & peach \\
         \midrule
         Supervised Fintune & 74.4 & 43.8 & 77.0 & 79.0\\
         RL w. ManiSkill & \makecell{95.0\\\small{(↑.27)}} & \makecell{60.5\\\small{(↑.38)}} & \makecell{64.0\\\small{(↓.16)}} & \makecell{62.5\\\small{(↓.21)}}\\
         RL w. World Model & \makecell{93.1\\\small{(↑.25)}} & \makecell{64.6\\\small{(↑.47)}} & \makecell{85.9\\\small{(↑.11)}} & \makecell{84.5\\\small{(↑.06)}}\\
        \bottomrule
    \end{tabular}
\end{table}

\section{Experiments}
In this section, we begin by detailing our experimental datasets and implementation details. We then evaluate the overall performance of our training paradigm in in-domain, out-of-domain, and multi-task settings. 

\subsection{Experimental Settings}
\label{sec:experimental_settings}
\paragraph{Datasets}
In our experiment, we utilize ManiSkill~\cite{mu2021maniskill} as the physical simulator and collect a dataset to train the world model. Specifically, we select 20 objects, 2 containers, and 2 tables, thus generating 80 scenes. We also select 4 tasks as follows:
\begin{itemize}
    \item \textbf{Pick and Place (P.\&P.)}: Pick up the object with the gripper and place it into the container.
    \item \textbf{Push}: Push the object forward for a distance with the gripper.
    \item \textbf{Pull}: Pull the object back for a distance with the gripper.
    \item \textbf{Move to Aim (M.A.)}: Move the object to the target carpet with the gripper.
\end{itemize}
Then we collect 500 optimal and 500 suboptimal trajectories for each task in each scene, a total of more than 300k trajectories. We employ the MPLib motion planner~\cite{guomplib} to generate the trajectories. 
We define an optimal trajectory as one that only connects the essential waypoints required to complete the task, without any redundant movements. To enhance policy robustness and ensure broad coverage of the robot's workspace, we also generate suboptimal trajectories. These are created using several methods: 
\begin{itemize}
    \item Add unrelated intermediate waypoints with position and rotation noise to the optimal path to increase exploration.
    \item Interpolate new intermediate points along the trajectory.
    \item Apply small-range position and rotation disturbances near essential waypoints to capture diverse data during critical manipulation (e.g., grasping and placing).
\end{itemize}

Each trajectory contains a text instruction, an action sequence, and multi-view camera observations (a third-view camera and a wrist-view camera). We record the delta pose of the end-effector, relative to the robot's base coordinate system, as the action.

\begin{figure}
    \centering
    \includegraphics[width=1\linewidth]{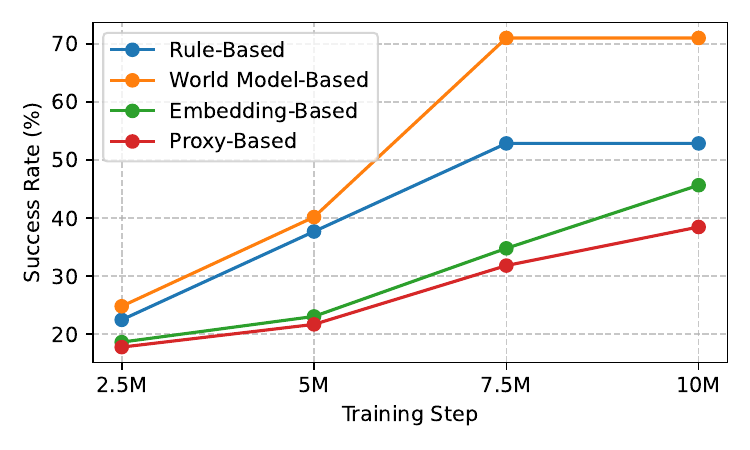}
    \vspace{-0.7cm}
    \caption{Success rate of policy trained with various rewards. We evaluate the SR for different policies with various reward modules before it converges.}
    \label{fig:reward_curve}
    \vspace{-0.5cm}
\end{figure}

\paragraph{Baselines}
We compare our paradigm training with the world model-based reward with three baselines:
\begin{itemize}
    \item \textbf{Rule-based reward}: Mainly utilized in physical simulators, which are calculated mainly based on environment states and a manually designed function.
    \item \textbf{iVideoGPT (Embedding-based reward)}~\cite{wu2024ivideogpt}: Directly use a single reward head from the hidden embedding to fit the manual reward labels.
    \item \textbf{DiWA (Proxy-based reward)}~\cite{chandra2025diwa}: Use an external model, practically an MLP in our experiment, to provide the reward with visual images.
\end{itemize}
For the rule-based reward, we conduct it in ManiSkill, and for the other two baselines, we incorporate the reward design as a reference and adapt it to our world model.
\begin{figure}
    \centering
    \includegraphics[width=1\linewidth]{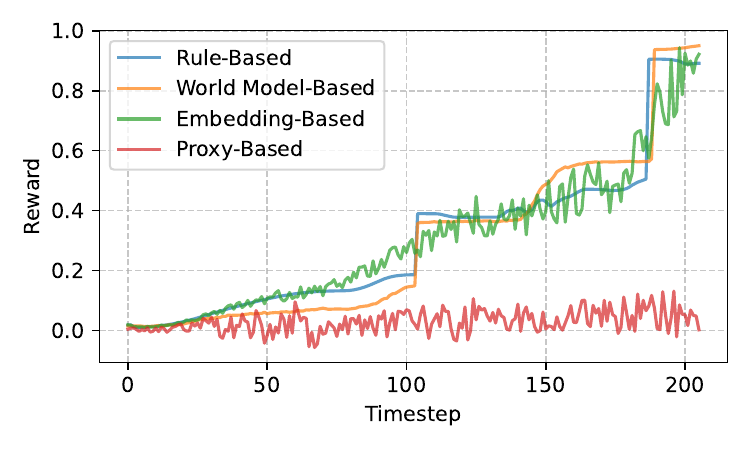}
    \vspace{-0.7cm}
    \caption{Reward curve of a successful trajectory in the out-of-domain environment for pick\&place task. It indicates that the world model-based reward is more generalizable to embedding-based and proxy-based rewards.}
    \label{fig:ood_reward}
    \vspace{-0.2cm}
\end{figure}
\begin{figure*}
    \centering
    \includegraphics[width=1\linewidth]{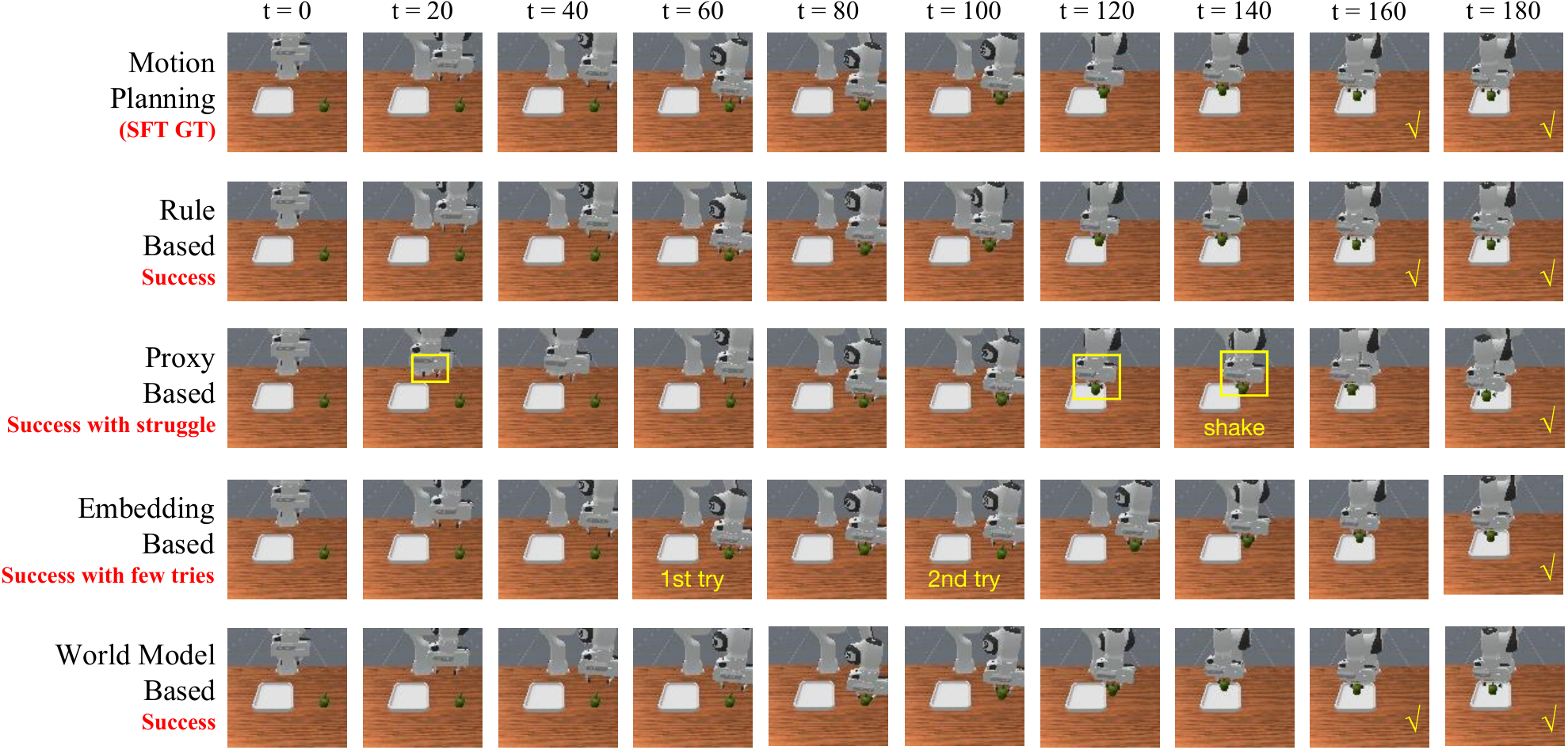}
    \caption{Visible cases for evaluation trajectories in the in-domain evaluations. The policy trained in the world model environment with general reward is comparable to those trained in physics simulators, while policies trained with proxy-based and embedding-based rewards exhibit inferior performance.}
    \vspace{-0.5cm}
    \label{fig:ind_vis}
\end{figure*}
\subsection{Implementation Details}
After collecting the dataset, we train our action and text world model using 80\% of the scenes, and for each scene, we use all the collected trajectories as training data. We preprocess the video into 28-frame clips with a frequency of 25 Hz, yielding approximately 51 million training clips. We train the world model for 5 epochs in approximately 96 hours on a cluster of NVIDIA-H20 GPUs. Then we conduct the RL training by encapsulating the world model-based environment as a new environment in the RL4VLA~\cite{liu2025can} framework.

We evaluate the Success Rate (\textbf{SR}) of the policy in the physical simulator. We evaluate the policy in the trained 80\% scenes for in-domain evaluation and in the 20\% unseen scenes for out-of-domain evaluation. In the out-of-domain environments, it contains unseen objects and unseen combinations between seen objects and seen containers.

\subsection{Overall Performance}
In order to evaluate the performance of the proposed training paradigm, we train the policy in both the world model and the physical simulator and evaluate the trained policy in the simulator. Specifically, we select MLP and OpenVLA as policies. We first use the data collected from the action planner to carry out the policy SFT. Then we use PPO~\cite{schulman2017proximal} to optimize MLP and OpenVLA-7B~\cite{kim2024openvla} in both the ManiSkill and our world model within 4 tasks separately.
Furthermore, we evaluate both the in-domain and out-of-domain ability of the policy. 
Then, we train the policy in a multi-task paradigm in P.\&P. and M.A. tasks, and evaluate the success rate to analyze how the world model benefits a task-level generalization policy training.

\paragraph{Comparasion with Baselines}
We evaluate the training time and the success rate of training policies with different reward modules, and the result is shown in Fig.~\ref{fig:reward_curve} by training an OpenVLA on the pick and place task, and evaluating the success rate of the trained policy. Our proposed reward achieves the best success rate in training a generalizable policy, which benefits from interacting with more environments. As for the training time, the world model-based RL training nearly achieves the same training efficiency as the rule-based training. However, the exogenous-based reward converges more slowly, and the success rate is far poorer than the rule-based rewards. 
The poor performance of exogenous-based reward lies in that it directly fits the manually designed reward function and thus provides an unstable reward. Moreover, our world model-based reward is an unsupervised paradigm and provides a stable reward, which benefits the RL training.

\begin{figure*}
    \centering
    \includegraphics[width=1\linewidth]{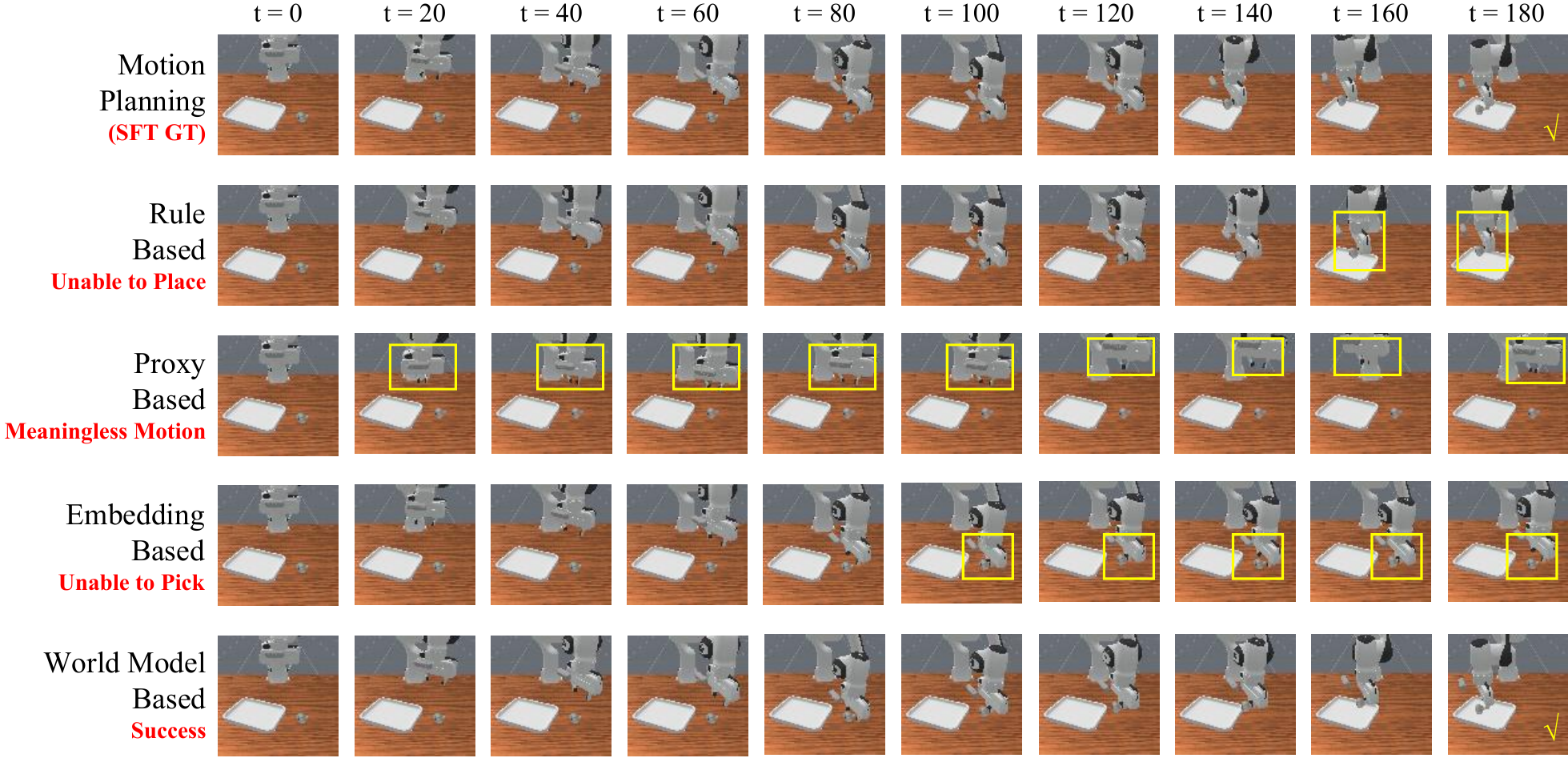}
    \caption{Visible cases for evaluation trajectories in the out-of-domain evaluations. Only the policy trained in the world model environment possesses generalization in OOD scenarios, while policies trained in other environments exhibit poor generalization since they only interact with a single environment.}
    \label{fig:ood_vis}
    \vspace{-0.3cm}
\end{figure*}

\paragraph{In-Domain Evaluation}
For in-domain evaluation, we train the policy in the initial scene and evaluate it in the same scene but with different environments. The results shown in Table~\ref{tab:overall} indicate that conducting RL in both ManiSkill and the world model can achieve a performance improvement, which indicates that the world model is able to serve as an effective environment for RL training. 

\paragraph{Out-of-Domain Evaluation}
For out-of-domain evaluation, we train the policy in the initial scene and evaluate it in a new scene, but with the same task. The results indicate that training the policy using the world model can achieve a significant improvement in unseen scenes. This is because the policy can interact with not only the given environment but also a series of additional environments to enhance the policy's understanding and learning the dynamics for the skill, instead of fitting to some property of the environment.

\paragraph{Multi-Task Evaluation}
The above evaluations demonstrate the generalization ability in new scenes for a single task, and in this part, we evaluate how the world model benefits the multi-task training for a policy. Specifically, we SFT a policy in two scenes for both P.\&P. and M.A. tasks and train it with RL in ManiSkill and the world model. The results are reported in Table~\ref{tab:multi_task}. It shows that optimizing the policy in the physical simulator with two manually designed rewards leads to an unbalanced performance, that is, the SR in P.\&P. rises while M.A. drops. However, optimizing the policy with the world model-based general reward achieves a balanced arisen for both tasks. This is because a policy needs to interact with two separate environments with two ``unaligned'' rewards, while it only needs to interact with one unified and general reward with the world model. 

\subsection{Detailed Analysis of the Reward Module}
\paragraph{Generalization for the reward module.}
In this part, we evaluate whether our designed reward can work well in an out-of-domain environment. We display the reward curve of a success trajectory in the OOD environment in Fig.~\ref{fig:ood_reward}. It indicates that both the rule-based and our world model-based reward work well when applied to new scenes, while the embedding-based reward performs worse, and the proxy-based reward performs the worst. This is because the rule-based reward is determined by the physical state of the simulator, which is not affected by the objects. For the proxy-based reward, the reward module is not trained on the new scene, thus providing a meaningless reward. Owing to the fact that the world model is trained on massive data, it learns a general dynamic understanding and achieves better generalizability, thus producing general and robust rewards.

\paragraph{Visualization for trajectories of trained policies.}
In this part, we provide some visualization cases in both the in-domain and out-of-domain trajectory visualizations for policies trained with different reward modules. For the in-domain environment, we choose ``pick up the green bell pepper and place it on the plate'', and for the out-of-domain environment, we choose ``pick up the garlic and place it on the plate''. The visualizations is shown in Fig.~\ref{fig:ind_vis} and Fig.~\ref{fig:ood_vis}.

For the in-domain evaluation, the rule-based and our world model-based reward achieve the best performance, while there are some effector shaking or gripper misclosing for the policy trained with proxy-based reward, and some double tries for the embedding-based policy. For the OOD evaluation, only the policy trained with the world model can succeed in the task. By comparison, the policy with rule-based rewards fails to learn how to lower the effector and place the object in some cases, while the policy trained with embedding-based rewards fails to pick up the object.

\section{Conclusion and Future Works}
In this work, we introduce RoboScape-R, a novel RL training paradigm where the world model serves as the environment by intrinsically providing a general and robust reward signal. By leveraging the world model as a universal environment simulator, we are able to train policies that exhibit enhanced generalization capabilities across diverse scenarios. Extensive evaluations demonstrate that utilizing our world model-based intrinsic reward yields significantly more generalizable policies, notably outperforming existing exogenous reward designs. For future work, this paradigm holds promise for adaptation to more complex, real-world tasks, paving the way for a substantial reduction of the Sim2Real gap.
\bibliographystyle{IEEEtran}
\bibliography{main}
\clearpage
\setcounter{page}{1}
\maketitlesupplementary

\section{Broader Impacts}
Our world model, as a scalable environment framework, offers positive value for training robotic policies. The world model environment can interact with actions output by the policy while providing observations and rewards for the next frame. Such rewards are unified, which enhances the generalization capability of the policy. This type of reward is ``endogenous'', which is derived from the world model’s understanding of diverse tasks—to facilitate multi-task generalization learning of the policy.

\section{Limitations}
While our framework enables the world model to act as an environment for training generalizable policies, we acknowledge several limitations:(1) Our current framework lacks robust support for policy learning in long-horizon and complex tasks. As our world model adopts an autoregressive architecture, it can only achieve stable rollout within 300 frames when the window size is set to 48 frames. Exceeding this limit may lead to deteriorated quality and controllability of generated videos. This restricts our tasks to short-duration scenarios, meaning we cannot yet accommodate long-range, complex tasks such as folding clothes. (2) Our framework relies on the empirical assumption that the world model has fully learned the dynamic transitions of the real world. However, this assumption hinges on the fundamental performance of the world model itself.

\begin{figure}
    \centering
    \includegraphics[width=1\linewidth]{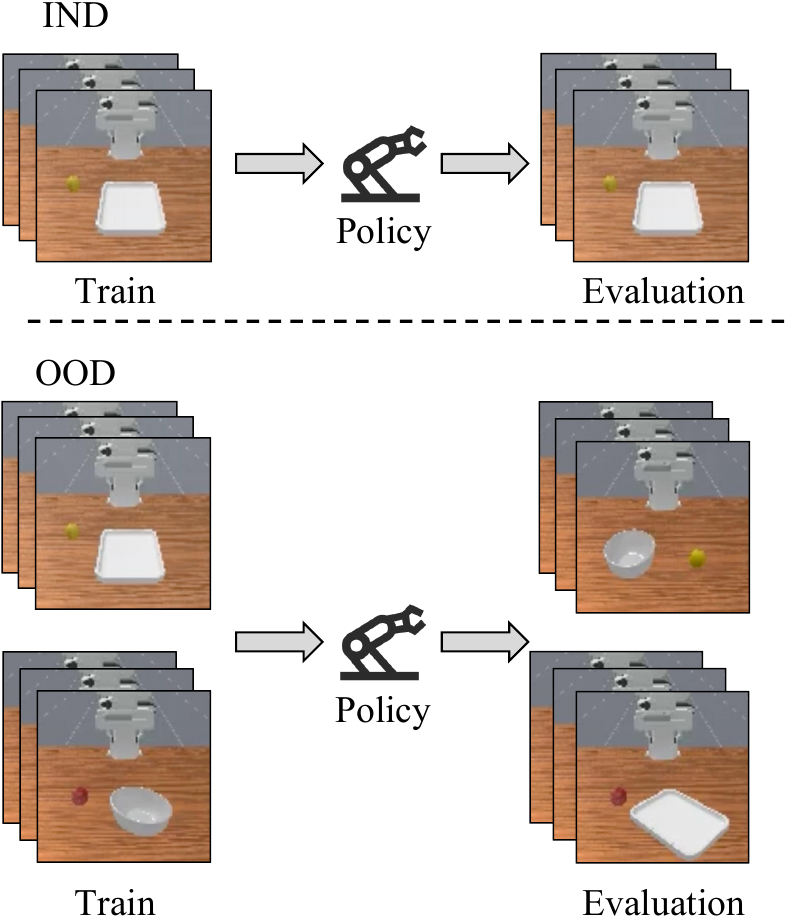}
    \caption{Task setting for in-domain and out-of-domain evaluation. For in-domain evaluation, training and evaluation are in the same environment, while for out-of-domain evaluation are in different environments.}
    \vspace{-0.5cm}
    \label{fig:train_setting}
\end{figure}
\section{Supplemented Evaluation Results}
\begin{figure*}[h]
    \centering
    \includegraphics[width=0.8\linewidth]{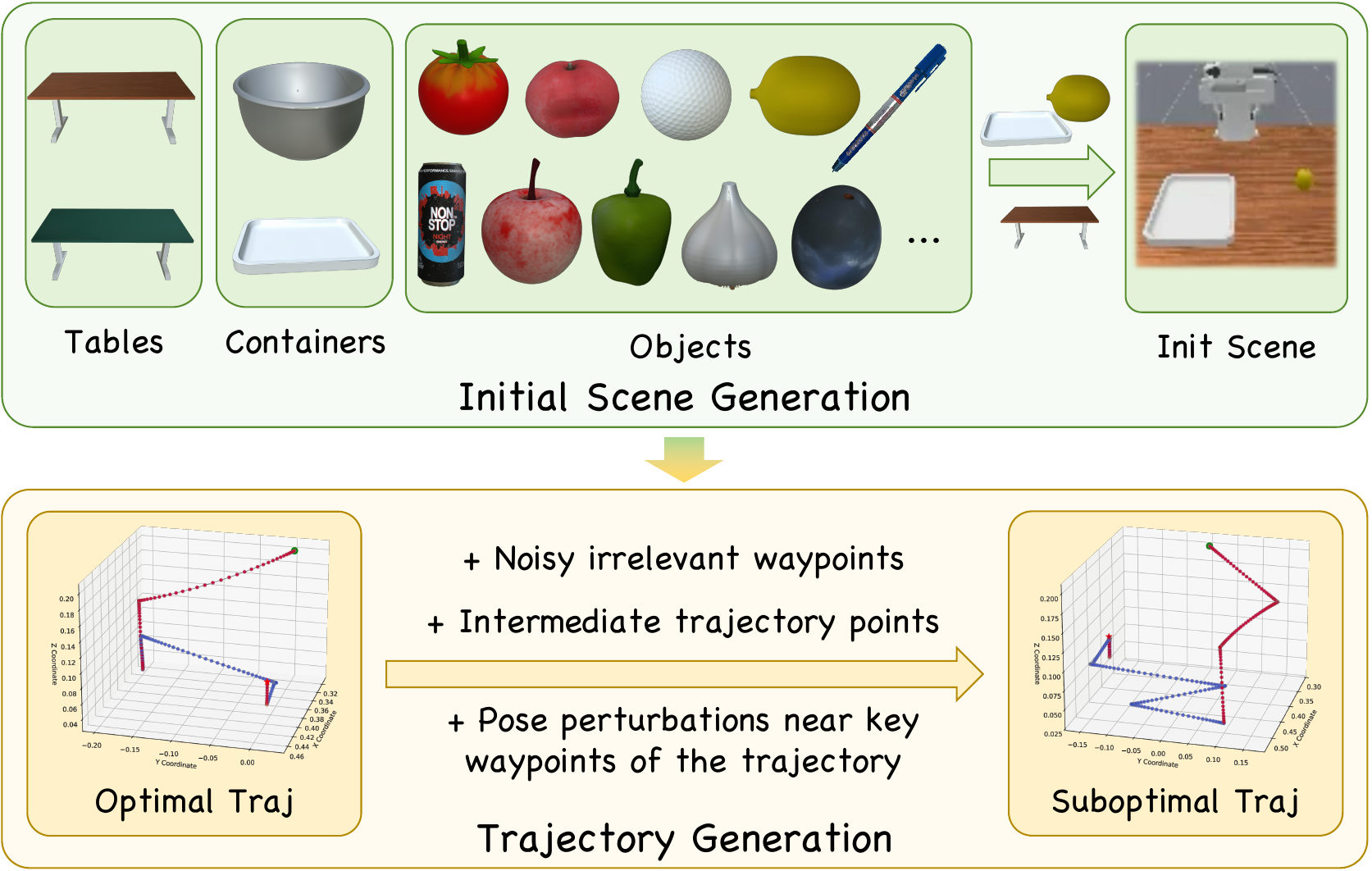}
    \caption{Diagram for the data collection pipeline. We first select tables, containers, and objects to create the initial scene, then we use motion planning to generate the optimal trajectory. We then modify the key waypoints and add the perturbation, and use motion planning to generate the suboptimal trajectories. }
    \label{fig:data_collection}
\end{figure*}
\subsection{Task Setting for In-domain and Out-of-domain Evaluation}

In the evaluation part, we conduct both the in-domain and out-of-domain evaluation. We display the task setting in Fig.~\ref{fig:train_setting}. For in-domain evaluation, we train the policy in one environment and evaluate it in the same environment but with different initial states. For out-of-domain evaluation, we train the policy in several environments and evaluate it in different environments with seen objects and containers, but with different combinations. For example, we train the policy in ``pick up the lemon and place it in the plate'' and ``pick up the peach and place it in the bowl'', and we evaluate the policy in ``pick up the lemon and place it in the bowl'' and ``pick up the peach and place it in the plate''.
\subsection{Supplemented Description for the Self-Collected Dataset}
\begin{figure*}[t]
    \centering
    \begin{subfigure}[b]{0.95\linewidth} 
        \centering
        \includegraphics[width=1\linewidth]{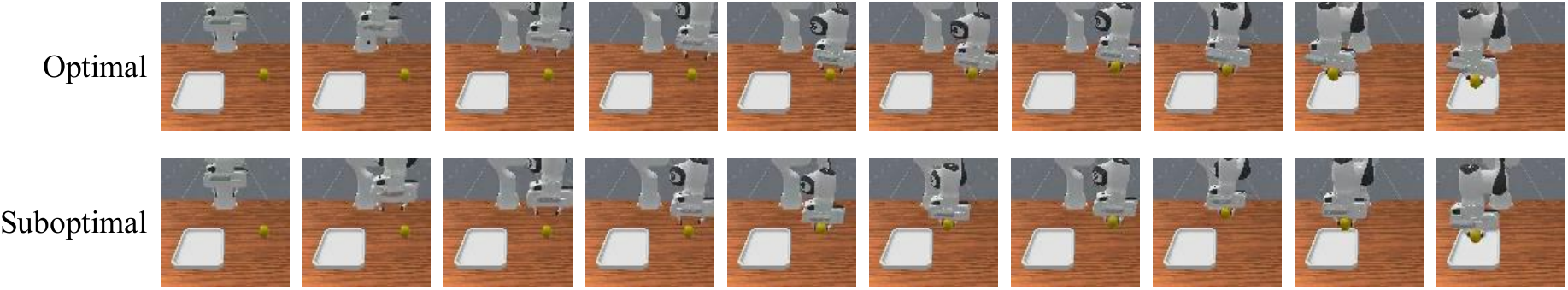} 
        \subcaption{Pick and Place (P.\&P.)} 
        \label{fig:trajectory_vis_1}
    \end{subfigure}
    \vspace{10pt} 
    
    \begin{subfigure}[b]{0.95\linewidth}
        \centering
        \includegraphics[width=1\linewidth]{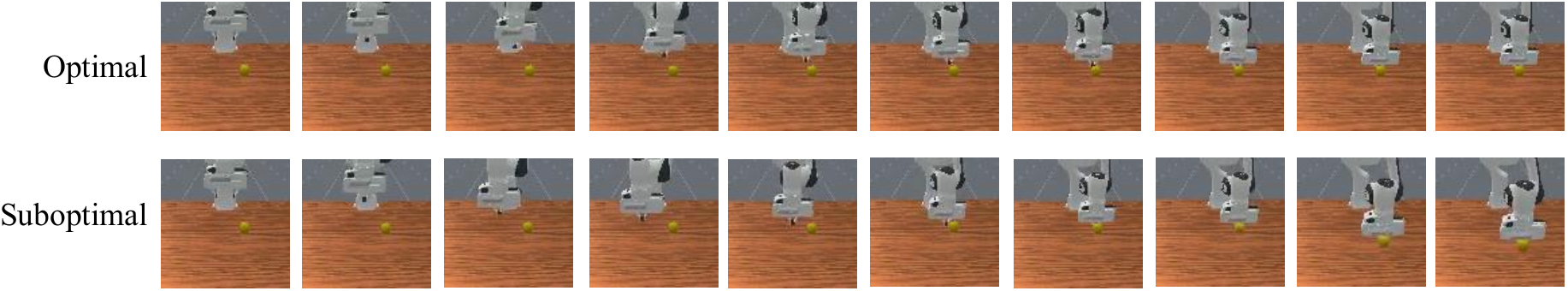
        } 
        \subcaption{Push}
        \label{fig:trajectory_vis_2}
    \end{subfigure}
    \vspace{10pt}
    
    \begin{subfigure}[b]{0.95\linewidth}
        \centering
        \includegraphics[width=1\linewidth]{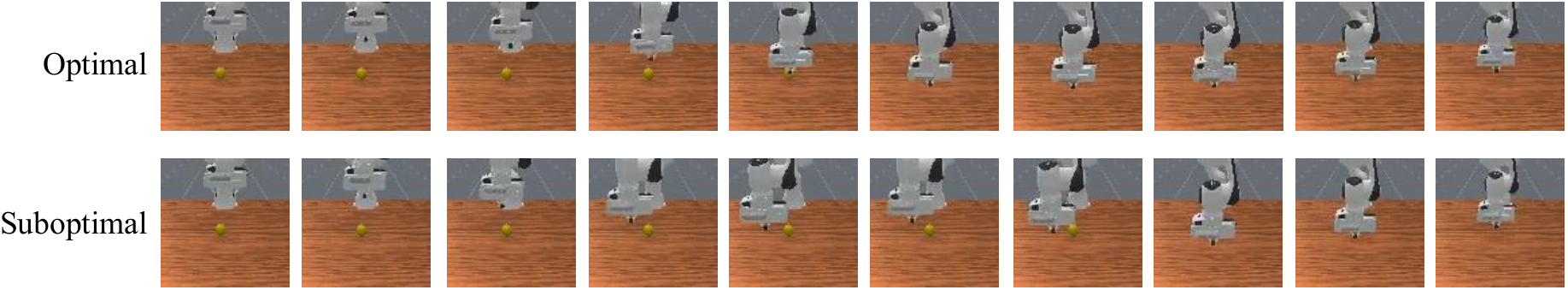} 
        \subcaption{Pull}
        \label{fig:trajectory_vis_3}
    \end{subfigure}
    \vspace{10pt}
    
    \begin{subfigure}[b]{0.95\linewidth}
        \centering
        \includegraphics[width=1\linewidth]{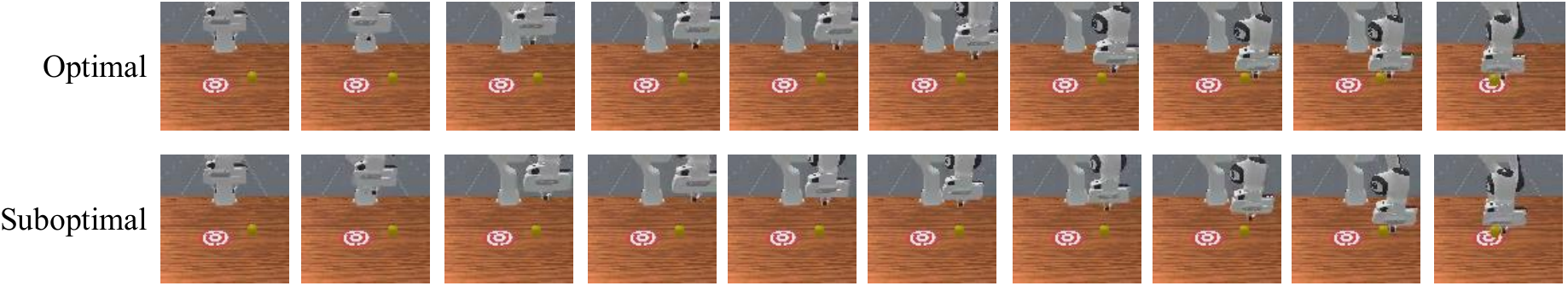} 
        \subcaption{Move to Aim (M.A.)}
        \label{fig:trajectory_vis_4}
    \end{subfigure}
    
    \caption{Visualization for optimal and suboptimal trajectories for different tasks.}
    \label{fig:trajectory_vis}
\end{figure*}

In our experiment, we have collected a dataset from ManiSkill~\cite{mu2021maniskill}. Specifically, we select 4 tasks, including pick and place, push, pull, and move to aim. For each task, we select 2 tables, 2 containers, and 20 objects to collect the data. A schematic diagram of data collection is presented in Fig.~\ref{fig:data_collection}.
In order to enable a more comprehensive learning of the action space for the world model to learning the dynamics, we also collect both optimal and suboptimal trajectories. The detail can be found in Sec.~\ref{sec:experimental_settings}, and we display the representative trajectories for each task in Fig.~\ref{fig:trajectory_vis}.

\subsection{Evaluation for the World Model Controlability}
\begin{figure*}
    \centering
    \includegraphics[width=1\linewidth]{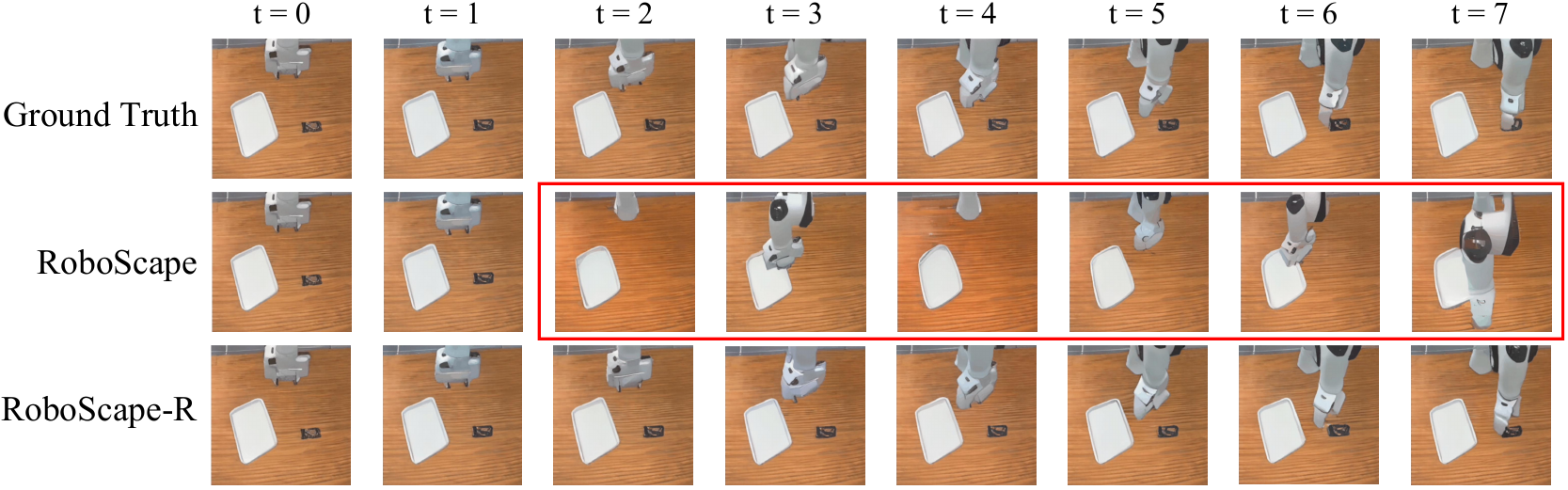}
    \caption{Visualization when the world model responds to an unseen trajectory in the out-of-domain environments.}
    \vspace{-0.5cm}
    \label{fig:bad_case}
\end{figure*}

Utilizing the world model as an RL environment also poses a challenge to the observation generation quality for the world model, mainly about the action controllability and the robustness to out-of-domain actions. This is due to that the world model is trained in a collected dataset, which indicates a discrete and limited action space, while the policy may generate an extreme action, especially at the beginning stage. As shown in Fig.~\ref{fig:bad_case}, our world model is able to respond to extreme actions due to the promoted cross-attention-based action injection and the comprehensive pre-training data, while other world models may suffer from meaningless observation generation due to the extreme action sequence.

\end{document}